\pdfoutput=1

\documentclass[11pt]{article}

\usepackage[preprint]{acl}

\usepackage{times}
\usepackage{latexsym}
\usepackage{svg}

\usepackage[T1]{fontenc}

\usepackage[utf8]{inputenc}

\usepackage{microtype}

\usepackage{inconsolata}

\usepackage{graphicx}
\usepackage{subfig}
\usepackage{amsmath}
\usepackage{tabularx}
\usepackage{booktabs} 
\usepackage{caption} 
\usepackage{multirow} 

\usepackage{xcolor}

\usepackage{todonotes}

\title{Towards Fundamental Language Models: Does Linguistic Competence Scale with Model Size?}

\author{
 \textbf{Jaime Collado-Montañez\textsuperscript{1}},
 \textbf{L. Alfonso Ureña-López\textsuperscript{1}},
 \textbf{Arturo Montejo-Ráez\textsuperscript{1}}
\\
\\
 \textsuperscript{1}Computer Science Department, University of Jaén, Campus Las Lagunillas s/n, Jaén, 23071, Spain
\\
 \small{
   \textbf{Correspondence:} \href{mailto:jcollado@ujaen.es}{jcollado@ujaen.es}
 }
}

\begin{document}
\maketitle
\begin{abstract}
Large Language Models offer impressive language capabilities but suffer from well-known limitations, including hallucinations, biases, privacy concerns, and high computational costs. These issues are largely driven by the combination of linguistic competence and factual memorization within a single monolithic model. This paper introduces and empirically supports the Fundamental Language Model (FLM) paradigm, which advocates for smaller, linguistically competent models that offload factual retrieval to external tools. We evaluate models ranging from 135M to 32B parameters across three dimensions: linguistic competence, external factual knowledge, and internal factual knowledge. Our findings reveal that while both linguistic competence and factual knowledge improve with scale, internal factual knowledge grows significantly faster, suggesting that model size is more closely tied to memorization than to core language ability. These results support a modular approach to language modeling, where compact, linguistically proficient models serve as the foundation for tool-augmented systems. The FLM paradigm offers a path toward more efficient, interpretable, and sustainable NLP solutions.
\end{abstract}

\section{Introduction}
\label{sec:sec1}
Large Language Models (LLMs) have demonstrated significant advancements in natural language processing, achieving state-of-the-art results across a range of natural language tasks. However, these models exhibit several limitations~\cite{bengio2025internationalaisafetyreport}, including the generation of inaccurate information (hallucinations), concerns regarding data privacy, and the propagation of biases present in the training data. A key factor contributing to these challenges is the substantial volume of factual information that LLMs internalize during training, where models simultaneously encode both factual data and linguistic structures.

This paper explores the viability of an alternative paradigm: Fundamental Language Models (FLMs). Instead of internalizing factual knowledge, FLMs aim to preserve the core linguistic competence of traditional LLMs while delegating factual retrieval to external knowledge sources. This separation could offer several advantages such as reducing model size, mitigating biases, and improving factual accuracy by relying on dynamically retrieved information rather than static, potentially outdated internalized knowledge.

In this context, linguistic competence is defined as the model's capacity to generate and comprehend language, demonstrating proficiency in linguistic structures (e.g., grammar, vocabulary, and meaning) independent of embedded factual data. Drawing on linguistic theory as defined by the Council of Europe in its Common European Framework of Reference for Languages (CEFR)\footnote{\url{https://www.coe.int/en/web/common-european-framework-reference-languages/cefr-and-its-language-versions}}, FLMs should prioritize three key sub-competences:

\begin{enumerate} 
    \item \textbf{Lexical competence}: Knowledge of, and ability to use, the vocabulary of a language consisting of word classes and fixed expressions \cite{marconi1997lexical}.
    \item \textbf{Grammatical competence}: Knowledge of, and ability to understand and express meaning by producing and recognising well-formed phrases and sentences \cite{MILLROOD2014259}.
    \item \textbf{Semantic competence}: The capacity to generate and understand meaningful phrases, sentences, and text, including resolving ambiguity, paraphrasing, and interpreting nuanced meanings in context \cite{marconi2020semantic}. 
\end{enumerate}

Phonological and orthoepic competences---related to spoken language---and orthographic competence—related to spelling—are less relevant to text-based models and thus remain outside the primary focus of FLMs.

While traditional retrieval-augmented generation (RAG) systems~\cite{lewis2020retrieval} enhance factual retrieval capabilities while maintaining full-scale language models, FLMs propose a more fundamental separation between linguistic and factual knowledge. Our research evaluates whether linguistic competence remains robust in smaller models as stated by~\citealt{eldan2023tinystories}, supporting FLMs as a viable direction for future development.

The paper is structured as follows: in Section \ref{sec:sec1}, we introduce FLMs and their potential to disentangle linguistic competence from factual knowledge, addressing critical challenges in current LLM architectures. Section \ref{sec:sec2} examines related work, reviewing transformer models, linguistic evaluation methods, and theoretical frameworks. Section \ref{sec:sec3} presents the evaluation methodology, detailing the assessment of linguistic competencies. Section \ref{sec:sec4} presents our experimental results, with findings in both linguistic competence and factual knowledge analysis. Finally, Section \ref{sec:sec5} synthesizes our results, demonstrating that linguistic competence stabilizes at smaller model sizes while factual knowledge continues to scale, supporting the viability of the FLM approach.

\section{Related Work}
\label{sec:sec2}
While transformer models exhibit impressive capabilities in handling linguistic tasks, they do not replicate traditional linguistic analysis methods. The models encode semantic roles and grammatical features in specific regions of sentence embeddings, rather than distributing this information evenly across the entire embedding \cite{nastase2024tracking}. Anyhow, it seems that each layer in a transformer captures different levels of linguistic information, from local to global dependencies \cite{garnier2024transformers}. This was already found in not-so-large models like BERT, and several works have found that linguistically related information is encoded in a hierarchical way, and some layers seem to focus on different aspects \cite{rogers2021primer}. From these findings, the next question that arises is: How large a language model has to be to become linguistically competent? Although larger models perform better, smaller models can still achieve significant results \cite{steuer2023large, eldan2023tinystories}. 

The evaluation of LLMs on linguistic competence is present in many benchmarks \cite{chang2024survey}. Some studies have focused on this type of test to identify the proficiency of language models in linguistic aspects like grammar, vocabulary, or syntax, and compared it to reasoning capabilities \cite{atox2024evaluating}. The work by \cite{dentella2024languagevivovssilico} examines LLMs' ability to understand uncommon meanings of common words, finding that even advanced models like GPT-4 perform worse than teenagers at this task. This reveals important limitations in LLMs' semantic understanding capabilities, despite their otherwise impressive language abilities. This could suggest that going to larger models may not scale up linguistic capabilities. 

The BabyLM challenge has been engaging the research community to train language models on a limited set of texts, with the aim to emulate the way humans learn in their infancy \cite{hu2024findingssecondbabylmchallenge}. One of the most interesting findings was that, even with such a constrained set of training material, the performance of the models was not too far from models trained over trillions of tokens, like LLaMA-2. Effective approaches included preprocessing of the training data and some enhancements to the transformer architecture. 

The Sapir-Whorf hyphothesis, also known as \emph{linguistic relativity}, proposes that language influences our understanding of the world and, even more, our cognitive skills \cite{penn2014linguistic}. This hypothesis, which dates back to the middle of the 20th century, has been partially supported by the ``emergent'' abilities of large language models, though it is still an open discussion \cite{schaeffer2023emergent}. Studies with pre-linguistic infants have shown abilities to understand physical causality and object permanence \cite{hespos2004conceptual}. Early research work on chimpanzees showed they could solve complex puzzles and understand cause-and-effect relationships without linguistic abilities \cite{premack1959toward}. We could conclude that reasoning is something more than language, as symbolic reasoning can occur without ``talking'' to ourselves. 

The rise of the so-called \emph{Agentic AI} paradigm has driven the evolution of artificial intelligence systems far from monolithic approaches \cite{acharya2025agentic}. So larger is not necessarily better, and the cooperation of several language models, with differentiated roles, is a promising path \cite{feng2025one}. Recent research has found that LLMs may have reached the peak in reasoning capabilities despite their size \cite{lin2025zebralogic}.

Linguistic relativity may not be fully right. Yet language strongly influences thought \cite{hypothesis_2022}. Large language models are still at the core of the most advanced solutions in artificial intelligence. If language by itself could be such a powerful tool in natural thinking, the pursuit of linguistic competence isolated from factual knowledge is justified \cite{liu2024ddkdistillingdomainknowledge}, as demonstrated by the rapid adoption of RAG tools. To the best of our knowledge, no prior study exists focusing on the trade-off between model size and linguistic competence, due to the variety of architectures and training objectives across available models.

\section{Evaluating Competencies}
\label{sec:sec3}
Fundamental Language Models aim to separate linguistic competence from factual knowledge, ensuring that models retain strong language-processing abilities while externalizing factual retrieval. To explore this hypothesis, we assess linguistic competence and factual knowledge performance across various model families, including SmolLM2~\cite{allal2025smollm2smolgoesbig}, Qwen2.5~\cite{qwen2.5}, Llama-3~\cite{grattafiori2024llama3herdmodels}, OLMo-2~\cite{olmo20242olmo2furious}, Falcon3~\cite{Falcon3}, Gemma-2~\cite{gemma_2024}, and Yi-1.5~\cite{ai2025yiopenfoundationmodels}, with model sizes ranging from 135M to 32B parameters. Our evaluation leverages well-established benchmarks within the LM Evaluation Harness \cite{eval-harness}---a unified framework to test generative language models--- to analyze different competencies in a zero-shot setting.

For each benchmark, we follow the standard evaluation protocols implemented within the LM Evaluation Harness. This typically involves generating responses from the models given the input prompts or contexts and then comparing these responses against the ground truth using the appropriate evaluation metrics (e.g., accuracy for classification tasks, F1-score and exact match for question answering, and BLEU/ROUGE scores for generative tasks like TruthfulQA).

According to our definition of FLMs, these models should excel in language-related tasks while struggling with factual knowledge tasks unless supplemented with external retrieval mechanisms. However, defining strict boundaries between linguistic competence and factual knowledge is challenging, as effective communication often relies on shared world knowledge. Despite this complexity, we focus on benchmarks that best capture these distinct abilities.

\subsection{Linguistic competence}
Linguistic competence, defined here as the model's capacity to generate and comprehend language with proficiency in linguistic structures, is a crucial aspect of evaluating the efficacy of FLMs. Following the framework of the CEFR, we focus on three key sub-competences: lexical, grammatical, and semantic competence. The goal is to select benchmarks that effectively assess these sub-competencies while minimizing the influence of external factual knowledge.

\subsubsection{Lexical competence}
Lexical competence refers to the knowledge and effective use of vocabulary. To evaluate this competence we use Word-in-Context (WiC)~\cite{pilehvar-camacho-collados-2019-wic}:

\begin{itemize}
    \item \textbf{WiC}: This dataset assesses word sense disambiguation, a fundamental aspect of lexical competence as defined by the CEFR, by presenting sentence pairs containing the same word. The task is to determine whether the word is used with the same meaning in both contexts. WiC directly aligns with the CEFR's emphasis on vocabulary in context and requires models to understand subtle differences in word usage.
\end{itemize}

\subsubsection{Grammatical competence}
Grammatical competence encompasses the knowledge and production of well-formed phrases and sentences, adhering to the rules of syntax and morphology. To this end, we find that the Benchmark of Linguistic Minimap Pairs (BLiMP) \cite{warstadt2020blimp} evaluates this competence in depth:

\begin{itemize}
    \item \textbf{BLiMP}: This benchmark consists of minimal sentence pairs designed to isolate specific grammatical phenomena. One sentence in each pair is grammatically correct, while the other contains a syntactic violation. BLiMP provides a rigorous evaluation of grammatical competence, testing the model's understanding of linguistic rules such as agreement, negation, and binding dependencies.
\end{itemize}

\subsubsection{Semantic competence}
Semantic competence concerns the model’s ability to generate and comprehend meaningful phrases and sentences, which includes understanding sentence-level meaning, resolving ambiguity, and recognizing nuanced language use. Benchmarks from the LM Evaluation Harness that assess this competence include Recognizing Textual Entailment (RTE) \cite{Dagan2005ThePR}, Multi-Genre Natural Language Inference (MNLI) \cite{N18-1101}, and Quora Question Pairs (QQP)\footnote{https://quoradata.quora.com/First-Quora-Dataset-Release-Question-Pairs}:

\begin{itemize}
    \item \textbf{RTE}: This benchmark measures whether a model can determine if one sentence logically follows from another. RTE requires deep semantic understanding to grasp the meaning of sentences and their logical relationships, aligning with the CEFR's focus on interpreting nuanced meaning in context.
    \item \textbf{MNLI}: This dataset tests a model's ability to classify sentence pairs as entailment, contradiction, or neutral, evaluating its capacity to capture meaning across different domains and contexts. 
    \item \textbf{QQP}: This task involves determining whether two questions are semantically equivalent. It tests the model’s ability to understand paraphrases and sentence-level meaning, making it a key benchmark for evaluating semantic competence.
\end{itemize}

These benchmarks collectively assess a range of semantic abilities crucial for effective language understanding, as defined within the CEFR framework.

\subsection{Factual knowledge}

We categorize factual knowledge into two types: external factual knowledge, which involves reasoning over provided information, and internal factual knowledge, which assesses the model’s memorization of factual data. This distinction is critical for evaluating FLMs, as they are designed to prioritize information extraction from external sources while minimizing reliance on memorized facts.

\subsubsection{External factual knowledge}
External factual knowledge refers to the ability of a model to utilize and reason with information provided in a given context. We evaluate this using datasets that provide a source passage or context to retrieve the answer from such as LAnguage Modeling Broadened to Account for Discourse Aspects (LAMBADA) \cite{paperno2016lambadadatasetwordprediction}, BoolQ \cite{clark2019boolq}, Choice of Plausible Alternatives (COPA) \cite{Gordon2011ChoiceOP}, Multi-Sentence Reading Comprehension (MultiRC) \cite{MultiRC2018}, and Reading Comprehension with Commonsense Reasoning Dataset (ReCoRD) \cite{record2018}:

\begin{itemize}
    \item \textbf{LAMBADA}: LAMBADA standard is a collection of narrative passages sharing the characteristic that human subjects are able to guess their last word if they are exposed to the whole passage, but not if they only see the last sentence preceding the target word.
    \item \textbf{BoolQ}: It is a question-answering dataset for yes/no questions where each example is a triplet of (question, passage, answer).
    \item \textbf{COPA}: This dataset assesses causal reasoning by presenting a premise and two alternative completions, requiring the model to select the most plausible one.
    \item \textbf{MultiRC}: It is a dataset of short paragraphs and multi-sentence questions that can be answered from the content of the paragraph.
    \item \textbf{ReCoRD}: Consists of queries automatically generated from CNN/Daily Mail news articles. The answer to each query is a text span from a summarizing passage of the corresponding news. As the framework provides F1-score and exact match metrics for this benchmark, we compute their average as the final score.
\end{itemize}

\subsubsection{Internal factual knowledge}
Internal factual knowledge refers to the factual information that the model has memorized during its training process. This knowledge is particularly relevant for traditional LLMs, which internalize vast amounts of data, however, the goal for FLMs is to minimize reliance on internalized facts, retrieving information dynamically from external sources. To evaluate internal factual knowledge, we use benchmarks that test the model's ability to recall specific facts without access to external context. These include TriviaQA \cite{JoshiTriviaQA2017} and TruthfulQA \cite{lin-etal-2022-truthfulqa}:
\begin{itemize}
    \item \textbf{TriviaQA}: It is a large-scale reading comprehension dataset that includes question-answer pairs authored by trivia enthusiasts. This dataset provides evidence documents automatically gathered that do not guarantee to contain all facts needed to answer the question. Consequently, the LM Evaluation Harness excludes these documents during evaluation, making TriviaQA a suitable benchmark for assessing internal factual knowledge.
    \item \textbf{TruthfulQA}: This benchmark is designed to test a model’s ability to generate factually accurate responses while avoiding common misconceptions. It comprises three tasks: (1) TruthfulQA Generation, where the model generates a 1-2 sentence response to a given question; (2) TruthfulQA MC1, a multiple-choice task requiring the selection of the single correct answer from 4-5 options; and (3) TruthfulQA MC2, which presents a question along with multiple true/false reference answers and scores the model based on the normalized probability assigned to the correct responses. For the first task, we compute the average between all accuracies provided by the framework, i.e., BLEU, ROUGE-1, ROUGE-2, and ROUGE-L.
\end{itemize}

\begin{figure*}[!htb]
    \centering
    \includegraphics[width=\linewidth]{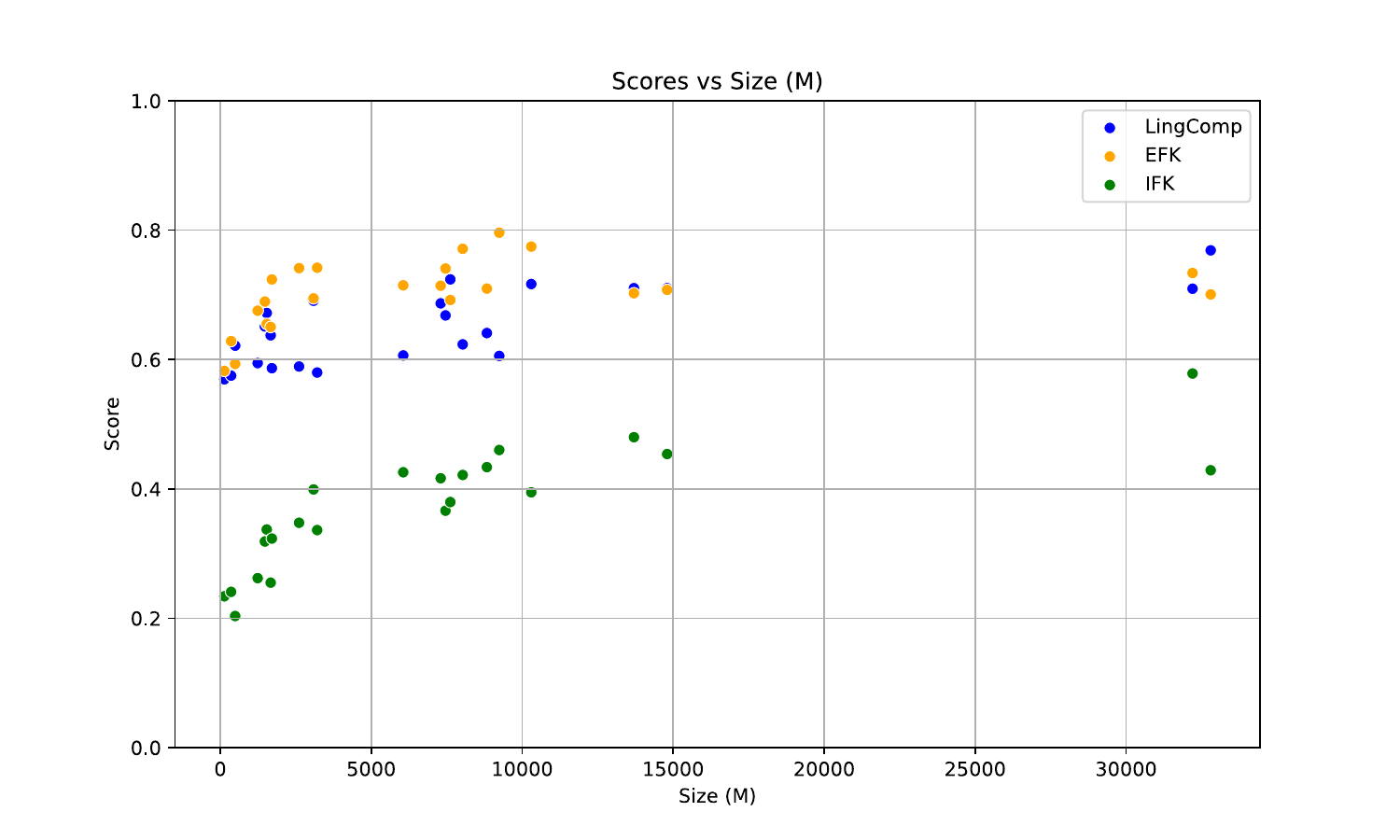}
    \caption{Scores achieved against model size in million parameters.}
    \label{fig:scores_vs_size}
\end{figure*}

\section{Results and discussion}
\label{sec:sec4}
In this section, we provide the results obtained for each competence separately. We also analyze them, including significance tests that prove that internal factual knowledge is more size-related than linguistic competence. Every experiment has been executed on two NVIDIA Ampere A100 GPUs.

Table \ref{tab:overall_results} presents the results of our evaluation across linguistic competence, external factual knowledge, and internal factual knowledge (see Figure~\ref{fig:scores_vs_size}). Overall, larger models achieve better results as expected, but we can see some model families stand out over others depending on the competence evaluated: Qwen2.5 models outperform similar-sized models on linguistic tasks, gemma-2 seems stronger in external factual knowledge-based benchmarks, and OLMo-2 does better in internal factual knowledge ones.

\begin{table*}[!htb]
    \centering
    \small
    \begin{tabular}{lcccc}
        \hline
        \textbf{Model} & \textbf{Size (B)} & \textbf{Ling. Comp.} & \textbf{External F.K.} & \textbf{Internal F.K.}\\
        \hline
        SmolLM2-135M & 0.135 & 0.5694 & 0.5824 & 0.2339 \\
        SmolLM2-360M & 0.36 & 0.5751 & 0.6285 & 0.2408 \\
        Qwen2.5-0.5B & 0.494 & 0.6214 & 0.5932 & 0.2034 \\
        Llama-3.2-1B & 1.24 & 0.5944 & 0.6755 & 0.2620 \\
        OLMo-2-0425-1B & 1.48 & 0.6511 & 0.6896 & 0.3187 \\
        Qwen2.5-1.5B & 1.54 & 0.6720 & 0.6550 & 0.3372 \\
        Falcon3-1B-Base & 1.67 & 0.6374 & 0.6503 & 0.2550 \\
        SmolLM2-1.7B & 1.71 & 0.5866 & 0.7238 & 0.3232 \\
        gemma-2-2b & 2.61 & 0.5892 & 0.7413 & 0.3476 \\
        Qwen2.5-3B & 3.09 & 0.6909 & 0.6946 & 0.3991 \\
        Llama-3.2-3B & 3.21 & 0.5799 & 0.7419 & 0.3363 \\
        Yi-1.5-6B & 6.06 & 0.6063 & 0.7147 & 0.4257 \\
        OLMo-2-1124-7B & 7.3 & 0.6868 & 0.7141 & 0.4164 \\
        Falcon3-7B-Base & 7.46 & 0.6682 & 0.7407 & 0.3663 \\
        Qwen2.5-7B & 7.62 & \underline{0.7239} & 0.6921 & 0.3796 \\
        Llama-3.1-8B & 8.03 & 0.6234 & 0.7712 & 0.4215 \\
        Yi-1.5-9B & 8.83 & 0.6409 & 0.7097 & 0.4336 \\
        gemma-2-9b & 9.24 & 0.6056 & \textbf{0.7961} & 0.4601 \\
        Falcon3-10B-Base & 10.3 & 0.7167 & \underline{0.7746} & 0.3947 \\
        OLMo-2-1124-13B & 13.7 & 0.7103 & 0.7025 & \underline{0.4799} \\
        Qwen2.5-14B & 14.8 & 0.7103 & 0.7077 & 0.4539 \\
        OLMo-2-0325-32B & 32.2 & 0.7095 & 0.7338 & \textbf{0.5784} \\
        Qwen2.5-32B & 32.8 & \textbf{0.7688} & 0.7007 & 0.4288 \\        
        \hline
    \end{tabular}
    \caption{Scores for each evaluated model and competence, ordered by model size. Competencies are computed as the average of all their selected tasks. Best scores are highlighted in \textbf{bold} and second-best scores are \underline{underlined}.}
    \label{tab:overall_results}
\end{table*}

\paragraph{Linguistic competence}
Table \ref{tab:linguistic_competence_results} presents the results for linguistic competence. Qwen2.5-32B achieved the highest overall linguistic competence score (0.7688), indicating a strong performance across lexical, grammatical, and semantic tasks. Qwen2.5-7B also performed competitively, scoring 0.7239, and even Qwen2.5-3B remains close to the largest models evaluated: 0.6909 as compared to OLMo-2-0325-32B, which scored 0.7095. These findings suggest that linguistic competence can remain stable even at moderate model sizes, depending more on architecture and training data decisions than on model size. Semantic competence results, which are an averaged score of several other tasks, are further described in Appendix \ref{sec:appendix}, Table \ref{tab:semantic_results}.

\paragraph{External factual knowledge}
Table \ref{tab:reasoning_results} presents the results for external factual knowledge evaluation. Gemma-2-9b achieved the highest score (0.7961), followed by Falcon3-10B-Base (0.7746), both outperforming larger models such as Qwen2.5-32B (0.7007). These results suggest that external factual knowledge, which is highly related to reasoning capabilities, does not continue to improve with model size after a certain threshold. These results are in line with the ones achieved by \citealt{lin2025zebralogic} and reinforce the idea that factual retrieval can be effective in modest-sized models.

\paragraph{Internal factual knowledge}
Table \ref{tab:factual_knowledge_results} reports the results for internal factual knowledge. OLMo-2-0325-32B demonstrated the highest internal factual knowledge score (0.5784), followed by its smaller version OLMo-2-1124-13B (0.4799). Distinct from the other two competencies, large models duplicated the scores obtained by small ones, which supports the hypothesis that internal factual knowledge is highly dependent on model size, as larger models tend to memorize more factual data.

\subsection{Analysis of results}

In order to better understand the effect of model sizes on the three different competences, we performed two different statistical analyses: (1) comparison of linear regression slopes and (2) significance tests.

\subsubsection{Linear regression slopes}

To model the relationship between capabilities and model scale more accurately, we use $\log(\text{Size})$ rather than raw parameter count. This transformation reflects widely observed empirical behavior in neural scaling laws, where performance tends to improve logarithmically with model size~\cite{kaplan2020scalinglawsneurallanguage}.

\begin{figure*}[htb]
    \centering
    \includegraphics[width=\linewidth]{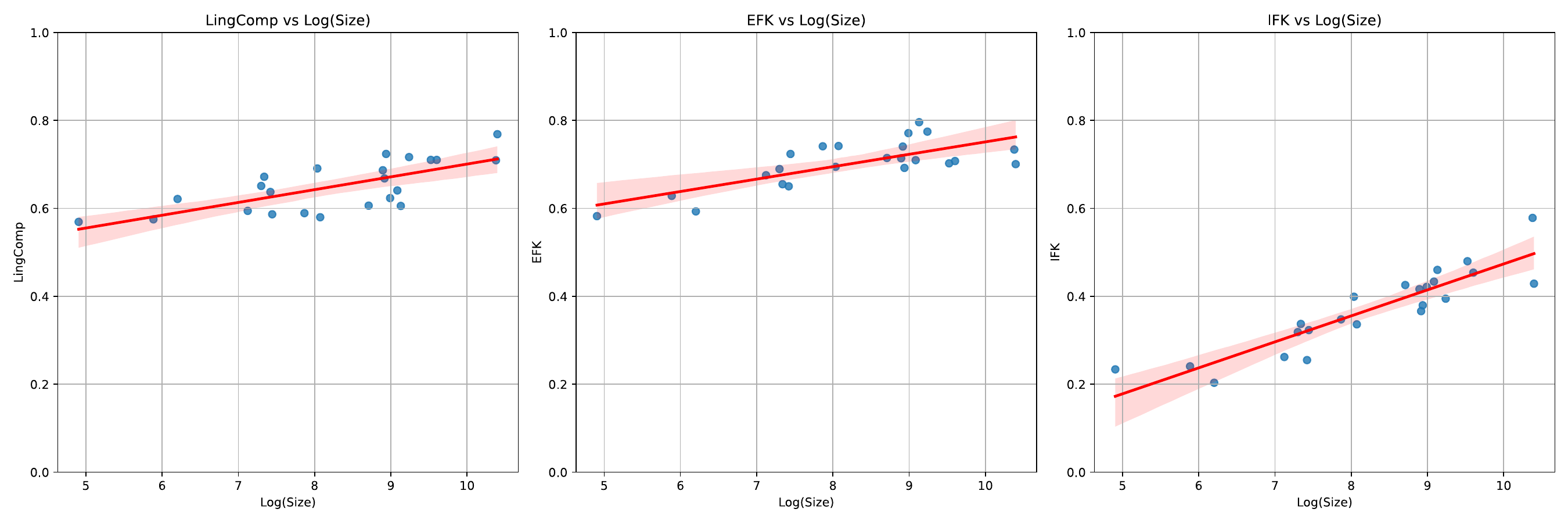}
    \caption{Linear regressions for each competence against $\log(\text{Size})$ in million parameters.}
    \label{fig:linear_regression}
\end{figure*}

A regression analysis using $\log(\text{Size})$ as predictor revealed that model size explains over 80\% of the variance in internal factual knowledge scores ($R^2 = 0.81$), but only about 50\% for linguistic competence. Additionally, the slope of the fitted regression line was more than twice as steep for internal factual knowledge ($0.059$) compared to linguistic competence ($0.029$), indicating that knowledge memorization grows significantly faster with model size than language proficiency (see Figure~\ref{fig:linear_regression}).

\subsubsection{Significance test}

To investigate the relationship between model size and performance, models were categorized based on the quartiles of their reported size. The first quartile (Q1) and the third quartile (Q3) of the `Size' variable were computed. Models were then assigned to one of three mutually exclusive categories:

\begin{itemize}
    \item Small (Q1): Models with a size less than or equal to the first quartile (size $\leq$ Q1).
    \item Medium: Models with a size greater than the first quartile but less than the third quartile (Q1 < size < Q3).
    \item Large (Q3): Models with a size greater than or equal to the third quartile (size $\geq$ Q3).
\end{itemize}

For each model, an average score was calculated for each of these three benchmark categories by taking the mean of the scores on the constituent individual benchmark tasks. Then, pairwise comparisons of average benchmark scores were conducted between these size categories to assess statistically significant differences in their distributions. Specifically, the following comparisons were performed:

\begin{itemize}
    \item Small (Q1) vs. Medium models
    \item Medium vs. Large (Q3) models
    \item Small (Q1) vs. Large (Q3) models
\end{itemize}

For each of these pairwise comparisons and for each of the three average benchmark scores (\textit{Avg LC}, \textit{ Avg EFK} and \textit{Avg IFK}), the Mann-Whitney U test was employed. The Mann-Whitney U test is a non-parametric test used to determine if there is a statistically significant difference between the distributions of two independent samples. The two samples for each test consisted of the collection of average scores for a given benchmark category belonging to the models within the two size categories being compared. A two-sided alternative hypothesis was used for all tests to detect if the distributions of scores differed in either direction. 

\begin{table*}[h!]
  \centering
  \small 
  \begin{tabularx}{\textwidth}{l *6{>{\centering\arraybackslash}X}}
    \hline
    \multirow{2}{*}{Benchmark Category} & \multicolumn{2}{c}{Small (Q1) vs Medium} & \multicolumn{2}{c}{Medium vs Large (Q3)} & \multicolumn{2}{c}{Small (Q1) vs Large (Q3)} \\
    \cmidrule(lr){2-3} \cmidrule(lr){4-5} \cmidrule(lr){6-7}
    & p-value & Significant & p-value & Significant & p-value & Significant \\
    \midrule
    External Factual Knowledge & 0.001 & Yes & 0.591 & No & 0.002 & Yes \\
    Internal Factual Knowledge & 0.015 & Yes & 0.078 & No & 0.009 & Yes \\
    Linguistic Competence & 0.256 & No & 0.062 & No & 0.015 & Yes \\
    \bottomrule
  \end{tabularx}
    \caption{Mann-Whitney U Test Results Summary (p-values and Significance at $\alpha=0.05$)}
    \label{tab:mannwhitneyu_results_acl}
\end{table*}

The results of the Mann-Whitney U tests (U statistic and p-value) are reported in Table \ref{tab:mannwhitneyu_results_acl}, and statistical significance was determined using a significance level (alpha) of 0.05.

The analysis using Mann-Whitney U tests suggests that statistically significant improvements in performance across all three benchmark categories (Linguistic Competence, External Factual Knowledge, and Internal Factual Knowledge) are most consistently observed when comparing the smallest models (first quartile) to larger models (either medium or third quartile). The step from medium-sized models to the largest models (third quartile) does not show statistically significant gains in any of the three benchmark categories based on these tests at the 0.05 significance level. This might indicate diminishing returns on performance with increasing model size beyond a certain point for these specific benchmarks and size categories.

\subsection{Discussion}

Our findings provide empirical evidence that linguistic competence---as measured across lexical, grammatical, and semantic subcompetencies---does not scale with model size as steeply as factual knowledge. While internal factual knowledge shows clear improvements with larger models, linguistic competence tends to stabilize at moderate model sizes, with diminishing returns beyond 5–7 billion parameters. In fact, if we group models by size with median size as split threshold into small and large models and calculate their average score for each competence, we find that IFK shows a 39.50\% performance difference between large and small models, while Linguistic Competence and EFK only shows 18.29\% and 8.47\% respectively.

These results reinforce the central hypothesis of the FLM paradigm: it is possible to decouple linguistic competence from factual knowledge, enabling the development of smaller models that retain strong language-processing abilities while delegating factual reasoning and retrieval to external systems. This approach could lead to more efficient, interpretable, and environmentally sustainable models by reducing parameter count without sacrificing linguistic performance.

These insights support a modular perspective on language model design, where factual retrieval, reasoning, and linguistic competence are treated as separable components. This not only aligns with cognitive models of language use but also opens the door to more flexible and specialized systems, where smaller FLMs serve as the linguistic interface while knowledge-intensive components are handled through dynamic retrieval.

\section{Conclusions}
\label{sec:sec5}
This paper investigated the viability of the Fundamental Language Model (FLM) paradigm, which proposes a modular approach to language modeling by decoupling linguistic competence from factual knowledge. Our empirical evaluation across a diverse set of language models and benchmarks revealed a consistent pattern: linguistic competence---encompassing lexical, grammatical, and semantic skills---tends to stabilize at moderate model sizes, whereas factual knowledge, particularly internalized facts, continues to scale with model size.

These findings have several important implications. First, they validate the core assumption behind FLMs---that strong language understanding capabilities do not require extremely large parameter counts or the memorization of vast factual corpora. Second, they support the design of more efficient, interpretable, and sustainable systems that delegate factual retrieval to external tools while relying on compact models for linguistic processing.

Our results also provide a quantitative basis for adopting a modular, tool-integrated architecture in future NLP systems. By treating linguistic competence as a standalone capability, developers can focus on optimizing each component of a language system separately-leading to more flexible, specialized, and controllable AI applications.

In sum, this work highlights the promise of the FLM paradigm and encourages a shift away from monolithic scaling toward targeted, capability-aware model design. Future research should extend this evaluation across languages, domains, and hybrid architectures to further explore the boundaries and potential of fundamental language modeling.

\section*{Acknowledgments}


This work has been partially supported by projects CONSENSO (PID2021-122263OB-C21), MODERATES (TED2021-130145B-I00), SocialTOX (PDC2022-133146-C21) funded by Plan Nacional I+D+i from the Spanish Government, and by the scholarship (FPI-PRE2022-105603) from the Ministry of Science, Innovation and Universities of the Spanish Government. Also, this work has been funded by the Ministerio para la Transformación Digital y de la Función Pública and Plan de Recuperación, Transformación y Resiliencia - Funded by EU – NextGenerationEU within the framework of the project Desarrollo Modelos ALIA.

\section*{Limitations}

While our research demonstrates the potential of Fundamental Language Models, several important limitations must be acknowledged. The separation of linguistic competence from factual knowledge presents challenges in cases where language understanding inherently requires world knowledge. For example, understanding metaphors, cultural references, or domain-specific terminology often depends on both linguistic and factual knowledge in ways that are difficult to disentangle. Our evaluation framework, though comprehensive, may not fully capture these interdependencies.

The performance stability we observed at smaller model sizes might not generalize across all linguistic tasks or languages. Our benchmarks focus primarily on English, and the relationship between model size and linguistic competence could vary significantly for other languages, particularly those with different syntactic structures or morphological complexity.

Our study also focuses on specific model architectures and sizes, and the findings might not extend to other architectural paradigms or scaling approaches. Future work should address these limitations through multilingual evaluation, real-world deployment testing, and investigation of hybrid approaches that better handle the linguistic-factual knowledge boundary.

\section{Ethical considerations}

The development of Fundamental Language Models raises some ethical considerations. While FLMs aim to reduce hallucinations and biases through external knowledge retrieval, this approach introduces new ethical issues. The selection and curation of external knowledge sources could perpetuate or amplify existing biases if not carefully managed. Additionally, the separation of linguistic and factual knowledge raises questions about transparency and accountability---users must understand which parts of the model's responses come from its linguistic processing versus external sources. Therefore, the separation between knowledge and linguistic competence does not ensure the avoidance of already existing problems in LLMs, but could help to identify and mitigate them.

\bibliography{custom}
\clearpage

\begin{onecolumn}
\appendix
\section{Appendix}
\label{sec:appendix}

\begin{table*}[!htb]
    \centering
    \small
    \begin{tabular}{l|ccc|c}
        \hline
        \textbf{Model} & \textbf{Lexical} & \textbf{Grammatical} & \textbf{Semantic} & \textbf{Linguistic} \\
        \hline
        SmolLM2-135M       & 0.5031    & 0.7911      & 0.4138    & 0.5694        \\
        SmolLM2-360M       & 0.4875    & 0.8050      & 0.4328    & 0.5751        \\
        Qwen2.5-0.5B        & 0.4937    & 0.8176      & 0.5528    & 0.6214        \\
        Llama-3.2-1B        & 0.4828    & 0.8246      & 0.4758    & 0.5944        \\
        OLMo-2-0425-1B      & 0.5408    & 0.8222      & 0.5902    & 0.6511        \\
        Qwen2.5-1.5B        & 0.5313    & 0.8251      & 0.6596    & 0.6720        \\
        Falcon3-1B-Base     & 0.5157    & 0.8240      & 0.5727    & 0.6374        \\
        SmolLM2-1.7B        & 0.5000    & 0.8024      & 0.4572    & 0.5866        \\
        gemma-2-2b          & 0.4937    & 0.7710      & 0.5028    & 0.5892        \\
        Qwen2.5-3B          & 0.6254    & 0.7270      & 0.7204    & 0.6909        \\
        Llama-3.2-3B        & 0.4969    & 0.8217      & 0.4212    & 0.5799        \\
        Yi-1.5-6B           & 0.5000    & 0.6936      & 0.6252    & 0.6063        \\
        OLMo-2-1124-7B      & 0.5235    & 0.8177      & 0.7190    & 0.6868        \\
        Falcon3-7B-Base     & 0.5533    & 0.8184      & 0.6330    & 0.6682        \\
        Qwen2.5-7B          & 0.5815    & 0.8225      & 0.7676    & \underline{0.7239}        \\
        Llama-3.1-8B        & 0.5110    & 0.8195      & 0.5398    & 0.6234        \\
        Yi-1.5-9B           & 0.6129    & 0.7054      & 0.6045    & 0.6409        \\
        gemma-2-9b          & 0.5125    & 0.7799      & 0.5244    & 0.6056        \\
        Falcon3-10B-Base    & \underline{0.6489}    & \underline{0.8289}      & 0.6723    & 0.7167        \\
        OLMo-2-1124-13B     & 0.5705    & 0.8100      & 0.7505    & 0.7103        \\
        Qwen2.5-14B         & 0.5188    & \textbf{0.8320}      & \underline{0.7801}    & 0.7103        \\
        OLMo-2-0325-32B     & 0.5737    & 0.8118      & 0.7429    & 0.7095        \\
        Qwen2.5-32B         & \textbf{0.6708}    & 0.8285      & \textbf{0.8071}    & \textbf{0.7688}        \\
        \hline
    \end{tabular}
    \caption{Averaged scores for each linguistic subcompetence. Linguistic competence is computed as the average between lexical, grammatical, and semantic scores. Best scores are highlighted in \textbf{bold} and second-best scores are \underline{underlined}.}
    \label{tab:linguistic_competence_results}
\end{table*}

\begin{table*}[!htb]
    \centering
    \small
    \begin{tabular}{l|ccc|c}
        \hline
        \textbf{Model} & \textbf{RTE} & \textbf{MNLI} & \textbf{QQP} & \textbf{Semantic Comp.}\\
        \hline
        SmolLM2-135M       & 0.4946    & 0.3399      & 0.4070    & 0.4138        \\
        SmolLM2-360M       & 0.5596    & 0.3522      & 0.3865    & 0.4328        \\
        Qwen2.5-0.5B        & 0.5884    & 0.3869      & 0.6831    & 0.5528        \\
        Llama-3.2-1B        & 0.5668    & 0.3585      & 0.5022    & 0.4758        \\
        OLMo-2-0425-1B      & 0.5126    & 0.4761      & 0.7819    & 0.5902        \\
        Qwen2.5-1.5B        & 0.7004    & 0.5254      & 0.7530    & 0.6596        \\
        Falcon3-1B-Base     & 0.6318    & 0.4540      & 0.6324    & 0.5727        \\
        SmolLM2-1.7B        & 0.5957    & 0.4075      & 0.3685    & 0.4572        \\
        gemma-2-2b          & 0.6137    & 0.4338      & 0.4610    & 0.5028        \\
        Qwen2.5-3B          & 0.7581    & 0.5505      & 0.8527    & 0.7204        \\
        Llama-3.2-3B        & 0.5451    & 0.3462      & 0.3722    & 0.4212        \\
        Yi-1.5-6B           & 0.7401    & 0.5437      & 0.5917    & 0.6252        \\
        OLMo-2-1124-7B      & 0.7076    & 0.5896      & 0.8599    & 0.7190        \\
        Falcon3-7B-Base     & 0.6390    & 0.5296      & 0.7303    & 0.6330        \\
        Qwen2.5-7B          & \textbf{0.8159}    & 0.6265      & 0.8605    & 0.7676        \\
        Llama-3.1-8B        & 0.6968    & 0.5084      & 0.4141    & 0.5398        \\
        Yi-1.5-9B           & 0.7834    & 0.5077      & 0.5224    & 0.6045        \\
        gemma-2-9b          & 0.6787    & 0.4849      & 0.4096    & 0.5244        \\
        Falcon3-10B-Base    & 0.7040    & 0.5426      & 0.7703    & 0.6723        \\
        OLMo-2-1124-13B     & 0.7256    & 0.6546      & 0.8713    & 0.7505        \\
        Qwen2.5-14B         & \underline{0.8014}    & \underline{0.6704}      & 0.8685    & \underline{0.7801}        \\
        OLMo-2-0325-32B     & 0.7329    & 0.6076      & \textbf{0.8882}    & 0.7429        \\
        Qwen2.5-32B         & \textbf{0.8159 }    & \textbf{0.7266}      & \underline{0.8788}    & \textbf{0.8071}        \\       
        \hline
    \end{tabular}
    \caption{Scores for each semantic competence task. Best scores are highlighted in \textbf{bold} and second-best scores are \underline{underlined}. Semantic competence score is computed as the average of these three benchmarks.}
    \label{tab:semantic_results}
\end{table*}

\begin{table*}[!htb]
    \centering
    \small
    \begin{tabular}{l|ccccc|c}
        \hline
        \textbf{Model} & \textbf{LAMBADA} & \textbf{BoolQ} & \textbf{COPA} & \textbf{MultiRC} & \textbf{ReCoRD} & \textbf{EFK} \\
        \hline
        SmolLM2-135M       & 0.3555    & 0.6031    & 0.6900    & 0.5565    & 0.7071    & 0.5824    \\
        SmolLM2-360M       & 0.4545    & 0.6180    & 0.7800    & 0.5058    & 0.7843    & 0.6285    \\
        Qwen2.5-0.5B        & 0.4349    & 0.6245    & 0.7400    & 0.3962    & 0.7704    & 0.5932    \\
        Llama-3.2-1B        & 0.5393    & 0.6404    & 0.7700    & 0.5670    & 0.8610    & 0.6755    \\
        OLMo-2-0425-1B      & 0.5672    & 0.6309    & 0.8300    & 0.5604    & 0.8592    & 0.6896    \\
        Qwen2.5-1.5B        & 0.5861    & 0.7291    & 0.8300    & 0.2857    & 0.8442    & 0.6550    \\
        Falcon3-1B-Base     & 0.4508    & 0.7147    & 0.7300    & \textbf{0.5720}    & 0.7839    & 0.6503    \\
        SmolLM2-1.7B        & 0.6257    & 0.7229    & 0.8300    & 0.5695    & 0.8711    & 0.7238    \\
        gemma-2-2b          & 0.6402    & 0.7343    & 0.8800    & 0.5588    & 0.8930    & 0.7413    \\
        Qwen2.5-3B          & 0.5905    & 0.7722    & 0.8500    & 0.3851    & 0.8752    & 0.6946    \\
        Llama-3.2-3B        & 0.6423    & 0.7339    & 0.8600    & \textbf{0.5720}    & 0.9012    & 0.7419    \\
        Yi-1.5-6B           & 0.6802    & 0.8034    & 0.8500    & 0.3426    & 0.8971    & 0.7147    \\
        OLMo-2-1124-7B      & 0.6829    & 0.8003    & 0.8900    & 0.2915    & 0.9061    & 0.7141    \\
        Falcon3-7B-Base     & 0.6336    & 0.8150    & 0.8900    & 0.4849    & 0.8799    & 0.7407    \\
        Qwen2.5-7B          & 0.6511    & 0.8468    & 0.9100    & 0.1588    & 0.8936    & 0.6921    \\
        Llama-3.1-8B        & 0.6738    & 0.8211    & 0.8700    & \textbf{0.5720}    & 0.9193    & 0.7712    \\
        Yi-1.5-9B           & 0.6990    & \underline{0.8584}    & 0.8900    & 0.1914    & 0.9095    & 0.7097    \\
        gemma-2-9b          & \underline{0.7231}    & 0.8398    & \textbf{0.9300}    & \underline{0.5668}    & \underline{0.9207}    & \textbf{0.7961}    \\
        Falcon3-10B-Base    & 0.6660    & 0.8223    & \textbf{0.9300}    & 0.5563    & 0.8984    & \underline{0.7746}    \\
        OLMo-2-1124-13B     & 0.7079    & 0.7388    & \underline{0.9200}    & 0.2290    & 0.9168    & 0.7025    \\
        Qwen2.5-14B         & 0.7091    & 0.8529    & 0.9000    & 0.1613    & 0.9152    & 0.7077    \\
        OLMo-2-0325-32B     & \textbf{0.7266}    & 0.8278    & \textbf{0.9300}    & 0.2628    & \textbf{0.9220}    & 0.7338    \\
        Qwen2.5-32B         & 0.7002    & \textbf{0.8719}    & 0.8800    & 0.1378    & 0.9138    & 0.7007    \\
        \hline
    \end{tabular}
    \caption{Scores for each external factual knowledge (EFK) task. It is computed as the average of all the selected tasks. Best scores are highlighted in \textbf{bold} and second-best scores are \underline{underlined}.}
    \label{tab:reasoning_results}
\end{table*}

\begin{table*}[!htb]
    \centering
    \small
    \begin{tabular}{l|cccc|c}
        \hline
        \textbf{Model} & \textbf{TriviaQA} & \textbf{TruthfulQA\_gen} & \textbf{TruthfulQA\_mc1} & \textbf{TruthfulQA\_mc2} & \textbf{IFK} \\
        \hline
        SmolLM2-135M       & 0.0512    & 0.2564    & 0.2399    & 0.3879    & 0.2339    \\
        SmolLM2-360M       & 0.1841    & 0.2332    & 0.2118    & 0.3343    & 0.2408    \\
        Qwen2.5-0.5B        & 0.1272    & 0.0358    & 0.2534    & 0.3973    & 0.2034    \\
        Llama-3.2-1B        & 0.2509    & 0.1888    & 0.2313    & 0.3768    & 0.2620    \\
        OLMo-2-0425-1B      & 0.3756    & 0.2996    & 0.2313    & 0.3684    & 0.3187    \\
        Qwen2.5-1.5B        & 0.2942    & 0.2876    & 0.3011    & 0.4661    & 0.3372    \\
        Falcon3-1B-Base     & 0.0303    & 0.2983    & 0.2656    & 0.4256    & 0.2550    \\
        SmolLM2-1.7B        & 0.3879    & 0.2861    & 0.2521    & 0.3667    & 0.3232    \\
        gemma-2-2b          & 0.5080    & 0.2800    & 0.2399    & 0.3624    & 0.3476    \\
        Qwen2.5-3B          & 0.4242    & 0.3644    & 0.3182    & 0.4894    & 0.3991    \\
        Llama-3.2-3B        & 0.5088    & 0.1943    & 0.2497    & 0.3922    & 0.3363    \\
        Yi-1.5-6B           & 0.4963    & 0.4685    & 0.2974    & 0.4405    & 0.4257    \\
        OLMo-2-1124-7B      & 0.6257    & 0.3192    & 0.2876    & 0.4333    & 0.4164    \\
        Falcon3-7B-Base     & 0.1111    & 0.4489    & 0.3721    & 0.5333    & 0.3663    \\
        Qwen2.5-7B          & 0.5038    & 0.0606    & \underline{0.3905}    & 0.5634    & 0.3796    \\
        Llama-3.1-8B        & 0.6170    & 0.3345    & 0.2827    & 0.4517    & 0.4215    \\
        Yi-1.5-9B           & 0.5447    & 0.4036    & 0.3195    & 0.4667    & 0.4336    \\
        gemma-2-9b          & \underline{0.6803}    & 0.4076    & 0.2987    & 0.4539    & 0.4601    \\
        Falcon3-10B-Base    & 0.1890    & \underline{0.5282}    & 0.3415    & 0.5201    & 0.3947    \\
        OLMo-2-1124-13B     & 0.6683    & 0.4097    & 0.3390    & 0.5026    & \underline{0.4799}    \\
        Qwen2.5-14B         & 0.5960    & 0.2338    & \textbf{0.4015}    & \textbf{0.5843}    & 0.4539    \\
        OLMo-2-0325-32B     & \textbf{0.7356}    & \textbf{0.6879}    & 0.3599    & 0.5303    & \textbf{0.5784}    \\
        Qwen2.5-32B         & 0.6171    & 0.1193    & \textbf{0.4015}    & \underline{0.5774}    & 0.4288    \\ \\
        \hline
    \end{tabular}
    \caption{Scores for each internal factual knowledge (IFK) task. It is computed as the average of all the selected tasks. Best scores are highlighted in \textbf{bold} and second-best scores are \underline{underlined}.}
    \label{tab:factual_knowledge_results}
\end{table*}
\end{onecolumn}
\end{document}